# Dynamical causality under invisible confounders

Jinling Yan[1,2,†], Shao-Wu Zhang[1,*], Chihao Zhang[3,4,†], Weitian Huang[5,6], Jifan Shi[7,8,9], Luonan Chen[2,6,10,*].

[1] MOE Key Laboratory of Information Fusion Technology, School of Automation, Northwestern Polytechnical University, Xi'an 710072, China

[2] Key Laboratory of Systems Biology, Shanghai Institute of Biochemistry and Cell Biology, Center for Excellence in Molecular Cell Science, Chinese Academy of Sciences, Shanghai 200031, China

[3] NCMIS, CEMS, RCSDS, Academy of Mathematics and Systems Science, Chinese Academy of Sciences, Beijing 100190, China

[4] School of Mathematics Sciences, University of Chinese Academy of Sciences, Beijing 100049, China

[5] School of Computer Science and Engineering, South China University of Technology, Guangzhou, Guangdong 510006, China

[6] Guangdong Institute of Intelligence Science and Technology, Hengqin, Zhuhai, Guangdong 519031, China

[7] Research Institute of Intelligent Complex Systems, Fudan University, Shanghai 200433, China

[8] State Key Laboratory of Medical Neurobiology and MOE Frontiers Center for Brain Science, Institutes of Brain Science, Fudan University, Shanghai 200032, China

[9] Shanghai Artificial Intelligence Laboratory, Shanghai 200232, China

[10] Key Laboratory of Systems Health Science of Zhejiang Province, School of Life Science, Hangzhou Institute for Advanced Study, University of Chinese Academy of Sciences, Chinese Academy of Sciences, Hangzhou 310024, China

[†] These authors contributed equally to this work

[*] Corresponding author: Luonan Chen; Shaowu Zhang

**Email:** lnchen@sibcb.ac.cn; zhangsw@nwpu.edu.cn

**Abstract**

Causality inference is prone to spurious causal interactions, due to the substantial confounders in a complex system. While many existing methods based on the statistical methods or dynamical methods attempt to address misidentification challenges, there remains a notable lack of effective methods to infer causality, in particular in the presence of invisible/unobservable confounders. As a result, accurately inferring causation with invisible confounders remains a largely unexplored and outstanding issue in data science and AI fields. In this work, we propose a method to overcome such challenges to infer dynamical causality under invisible confounders (CIC method) and further reconstruct the invisible confounders from time-series data by developing an orthogonal decomposition theorem in a delay embedding space. The core of our CIC method lies in its ability to decompose the observed variables not in their original space but in their delay embedding space into the common and private subspaces respectively, thereby quantifying causality between those variables both theoretically and computationally. This theoretical foundation ensures the causal detection for any high-dimensional system even with only two observed variables under many invisible confounders, which is actually a long-standing problem in the field. In addition to the invisible confounder problem, such a decomposition actually makes the intertwined variables separable in the embedding space, thus also solving the non-separability problem of causal inference. Extensive validation of the CIC method is carried out using various real datasets, and the experimental results demonstrates its effectiveness to reconstruct real biological networks even with unobserved confounders.

**Significance Statement**

Identification of causal interactions in a complex dynamical system is crucial yet challenging across various disciplines, especially in the presence of numerous invisible/unobservable confounders. A novel method is proposed to overcome the challenges to infer dynamical causality under invisible confounders and further reconstruct the invisible confounders from time-series data by developing an orthogonal decomposition theorem in a delay embedding space. A virtue of our method is that it can distinguish the relationships between variables, e.g., the actual causality or the common interaction caused by invisible confounders and reconstructing the invisible confounders. This advancement not only addresses a longstanding problem in the field but also holds great importance for contemporary AI learning and data science, which currently lack such solutions.

**Main Text**

**Introduction**

   Identification of causal interactions in a complex dynamical system is crucial yet challenging across various disciplines (1, 2), including biology, ecology and deep learning, especially in the presence of numerous invisible/unobservable confounders. In many real systems, the system details are typically unknown and only a number of the observations or measured variables are available. Consequently, developing an accurate and reliable data-based causal detection method with the unobserved confounders becomes imperative and pressing, which is a long-open problem in data science and deep learning fields.

Depending on the type of the measured data, there are mainly two approaches for causality inference: statistical methods for cross-sectional data and dynamical methods for time-series data (3, 4). The causality inference from a statistical perspective mainly holds that causality is contained in internal actions of random variables and can be obtained by statistical randomization or intervention, under the assumption of a time-independent steady-state system or cross-sectional data. The representative statistical methods of causality inference include the famous potential outcome model (POM) from Rubin (5), the structural causal model (SCM) proposed by Pearl (6-8), and many related algorithms, such as the famous Peter-Clark (PC) algorithm (9), Linear Non-Gaussian Acyclic Model (LiNGAM) (10), Optimal Structure Identification With Greedy Search (GES) (11), Max-Min Hill-Climbing Bayesian network (MMPC) (12), etc. On one hand, these methods are mainly applicable to time-independent data or intervention data, thus unable to take full advantage of the dynamical information of the time series data that is widely accessible. On the other hand, they dedicate to inferring causality based on known directed acyclic graphs (DAG), which are not applicable to many real systems with feedback loops, and they may also encounter Markov equivalence classes that cannot be distinguished. To overcome these limitations, researchers proposed many effective algorithms from a dynamical perspective to identify causality for time-series data, including Granger causality (GC) (13), mutual prediction method (14), state space method (15), quantifying information method (16), recurrence plots method (17), convergent cross method (CCM)(2), dynamical causality (DC) framework(3), cross map evaluation(CME) (18), cross map smoothness (CMS) (19), partial cross mapping (PCM) (1) and conditional cross-map-based techniques (4, 20, 21), etc. GC is the one of the famous measures to identify causality between different variables, however, GC is mainly suitable for linear causality. Transfer entropy (TE) extends GC to nonlinear cases based on information theory, but it still cannot deal with the non-separability problem of nonlinear dynamics. The core idea of GC or TE is to measure the causality by predicting one variable from another. In contrast, the reconstruction-based technique originated from the delay embedding theory (22), by which the state space reconstruction technique has been developed and widely applied in nonlinear time-series analysis, including CCM, CME, CMS, etc. These methods can successfully study pairwise causal associations by constructing a variable from another from time-series data. However, challenges arise when attempting to detect causality with unobservable variables.

Usually, there are many confounders between two variables in a complex system, also known as common drivers, which may cause fan-out spurious causality. The confounders can be divided into two categories, i.e., visible confounders and invisible confounders. The erroneous or spurious causality caused by visible/observable confounders can be eliminated by conditional causal methods, such as partial cross mapping (PCM), conditional Granger causality (cGC), direct CMC (DCMC), etc. However, these methods need to iterate through all variables of the whole system to accurately exclude spurious or false causality, leading to high computational cost and also instability. Especially, these methods are effective only when all confounders are observable or known. In practice, many of the confounders in a complex system are invisible and unknown, thus making the existing methods fail or easily lead to spurious causations. In other words, most existing methods cannot accurately detect causality under invisible confounders, let alone the reconstruction of the confounders. Therefore, the problem of accurately quantifying causation by eliminating the effects of invisible confounders for a general system is not fully studied and remains outstanding. It is in urgent need of developing a new method to detect causality even with many invisible confounders and to further reconstruct those invisible confounders.

To fill this gap, we develop an approach, Causality under Invisible Confounders (CIC), to detect dynamical causality from time series data in both theoretical and computational ways. This method aims to accurately detect causal associations and further reconstruct invisible confounders, only from the observed data. The core idea is to transform original variables into their delay-embedded counterparts, which are then decomposed into their common and private subspaces, respectively, thereby quantifying effect and cause among those variables. Our novel orthogonal decomposition theorem ensures this decomposition, making the intertwined variables separable in the embedding space and thus solving the non-separability problem. Note that effect variables reconstruct their causal variables in the delay embedding space in contrast to the original space where cause variables predict their effect variables. Computationally, we adopt deep neural networks to make such an orthogonal decomposition effectively, thereby enabling accurate detection on causal links merely from observed time-series data even in the presence of unobserved confounders.

In this causal framework, our orthogonal decomposition theorem provides a theoretical basis for ensuring dynamical causality, while deep learning serves as the backbone to computationally realize the variable decomposition and confounder reconstruction in a nonlinear system. The CIC method thus provides an elegant solution for the long-standing problem of misidentifying spurious causal influences even with unobserved confounders. Notably, the CIC method enables the determination of causal associations and confounders using just two observed variables, a feat that is typically impossible with any other methods. It also has an outstanding advantage over existing causality inference methods in that it effectively reconstructs unobserved confounders. The method is extensively validated through various benchmark systems and real datasets with different causal structures. The applications to various systems demonstrate its effectiveness as a powerful tool to analyze and reconstruct the real causal networks, solely using time series data of partially observed variables.

**Results**

In the following, we extensively validate the proposed CIC on various simulated and real-world datasets, including gene regulatory network, ecological network, and neuron network of rhesus monkey. Firstly, the effectiveness and robustness of CIC are validated by three benchmark systems: a three-dimensional system with different couplings, an eight-dimensional logistic system, and five 10-node and five 100-node gene regulatory networks from DREAM4 in silico Network Challenge (23-25). Then, four real-world datasets, including food chain network (26-28), Hong Kong air pollution and cardiovascular disease dataset (29), Monkey brain neural dataset (4), and Rat circadian rhythm gene expression data (30) are used to validate the method. The result of the eight-dimensional logistic system and Monkey brain neural dataset are provided in **Supplementary Text**. We compare the CIC with eight other popular causal methods, including GC, TE, CCM, CME, CMC, PCM, DCME and DCMC. Specifically, GC, TE, CCM, CME, CMC infer causality based on pairwise observed variables, similar to CIC, while PCM, DCME, DCMC require additional observed variables as conditional variables to detect causal influence. As there is no universal standard to quantify the accuracy of causal inference, here we used the widely used indices to evaluate causal inference, e.g., AUC/ROC, Accuracy, Precision. In addition, since conventional methods cannot reconstruct invisible confounders, then reconstructions of confounders by CIC are compared with the ground truth.

**Inferring causality and reconstructing invisible confounders in benchmark systems**
*Three-dimensional system with different couplings*

To validate the CIC method, we consider a three-variable logistic system:

$$\begin{cases} x_{t+1} = x_t[\gamma_x - \gamma_x x_t - \beta_{yx} y_t - \beta_{zx} z_t] + \epsilon_{x,t} \\ y_{t+1} = y_t[\gamma_y - \gamma_y y_t - \beta_{xy} x_t - \beta_{zy} z_t] + \epsilon_{y,t} \\ z_{t+1} = z_t[\gamma_z - \gamma_z z_t] + \epsilon_{z,t} \end{cases}$$

where $\gamma_x = 3.7, \gamma_y = 3.72, \gamma_z = 3.78$, $\beta_{yx}$, $\beta_{zx}$, $\beta_{xy}$, $\beta_{zy}$ are coupling parameters, and different combinations of coupling parameters can lead to different network motifs. By setting different coupling parameters, we focus on the causal association from $x$ to $y$ under the four cases, as shown in **Fig 1A**, including the causality from $x$ to $y$, the inverse causality from $y$ to $x$, the confounder of $x$ and $y$, and non-causality between $x$ and $y$. Our causal metrics, i.e., CIC, is able to accurately discriminate the associations in the four different situations. Note that CIC not only infer causality but also quantify confounder. Then, we investigate the robustness and effectiveness of CIC at different levels of parameter perturbation., e.g. different noises, coupling strength and length of time series. As shown in the left panel of **Fig 1B**, CIC successfully infer the casual associations under various noise settings. Notably, CIC can distinguish direct cause $x \to y$ (System1) from the confounder case (System3) with sufficiently large coupled strength and a long length of the observed time series with sorely relying on the observations of $x$ and $y$ (**Fig 1B**, right two panels), This capability is not achievable with other methods. The quantification of invisible confounders under different parameters is provided in **Fig 1C**. Similarly, the high correlations indicates that CIC accurately reconstruct the invisible confounder with sufficiently large coupled strength and long length of the observed time series (**Fig 1C**, the right two panels).

Specifically, in the first panel of **Fig 1B-C**, the coupling strength and data length are fixed at 0.35 and 5000, respectively, which were all generated from the equations above, and the noise level is from 0.001 to 0.015, beyond which the logistic system will no longer converge. As the noise increases, the CIC can still robustly infer causality and identify non-causal case. And for the System3 with a confounder, the correlation between the reconstructed confounder $z_{xy}$ and the real confounder $Z$ by CCA is asymptotically stabilized to 0.6, while the correlation values in other cases are close to 0 with noise fluctuating. In the second panel of **Fig 1B-C**, the noise and length are fixed at 0.001 and 5000, respectively, and the coupling strength is changed from 0 to 0.6, for which the CIC can also effectively infer causality and identify confounders as the strength increases, even under weak coupling or strong coupling. And the correlation between reconstructed confounder and real confounder stabilizes at 0.6. In the third panel of **Fig 1B-C**, the noise and strength are fixed at 0.001 and 0.35, respectively, and the length of time series varied from 100 to10000. We can see that the CIC works well when the data length is greater than 2000. The above result is consistent with our theoretical results. And the CIC score is above 0.75 when there is causality and increases with the causal strength. The CIC score fluctuates between 0.25 and 0.75 when there is a confounder of $x$ and $y$. For the non-causality case, the CIC score is significantly less than 0.25 even under various parameter perturbations.

To demonstrate the robustness and effectiveness of the method, the comparisons with other eight approaches under different noise, causal strength and length in System1 and System3 are provided. **Fig 1D** and **Fig 1E** illustrate the accuracy and precision of the nine methods, respectively. The results show that CIC outperforms these methods in various situations, e.g. with the noise level, coupled strength, and time-series length as 0.001, 0.35, and 5000, respectively. **Fig 1G** shows that CIC can accurately identify all causal associations, but there are false positive results for GC, CCM, CME, CMC, PCM, DCMC and false negative situations for TE and DCME. In addition, CIC method is capable of identifying confounders, which is consistent with the ground truth. **Fig 1F** illustrates the accuracy of CIC in identifying confounders under different parameters. Our method is robust for various noise level, coupling strength and data length, even with invisible confounders.

### *Gene regulatory networks from DREAM4 in silico Network Challenge*

We then applied CIC to infer gene regulatory networks (GRNs) collected from the DREAM4 *in silico* Network Challenge. The synthetic gene expression data, generated by Gene Net Weaver, and based upon patterns found in model organisms, provides the basis for inference (23, 24). There are ten GRNs with different structures, including five networks of size 10 and five networks of size 100. And there are 5 different time series for each network of size 10 and 10 time series for each network of size 100, with 21 time points in each time series. And the ground truth network was also distributed along the gene expression data, which allows us to assess the quality of inference result.

The inference result in causality and confounders by CIC and the comparison with other approaches are shown in **Fig 2** and **Supplementary Text 2.2**. **Fig 2** illustrates the results of our method in five 10-node networks. Taking Net1 as an example, the first panel of **Fig 2A** is the true network of Net1, and the second and third panels depict the comparison of the CIC and other methods with the gold standard. The ROC curves and Accuracy are used to exhibit the performance of the different methods. The results indicate that our method significantly outperforms other methods, both by ROC and Accuracy criterion. Moreover, our method has the unique advantage of quantifying confounders. The fourth panel of **Fig 2A** globally shows the correlation between the reconstructed confounders, i.e., the common latent variable, and the real confounders, which approaches to 0.7. While the correlation between the common latent variable of other non-confounder cases and the node with the lowest degree is less than 0.3. This shows that reconstructing confounders is of high quality and avoids the misidentifications for the non-confounder cases. In addition, the details of the reconstructed confounder are further analyzed. The distribution of high correlations also reflects the consistency of the reconstructed confounders marked with green boxes in the sixth panel and the real confounders in the ground truth matrix from the fifth panel.

The same analysis is also performed for the other nine networks in DREAM4 dataset. In general, CIC has a significant advantage over existing methods. For 10-node networks, the AUC of CIC is between 0.8-0.9, while the AUC of other methods is about 0.7. For the application of 100-node networks, the Accuracy criterion shows that our method significantly exceeds other methods in the causality inference of high-dimensional gene regulatory networks. The detailed result of five 100-node networks is provided in **Supplementary Text 2.2.** And the reconstruction of confounders reconstruction is more accuracy in the 100-node network. For example, in the highly correlated grids, i.e., reconstructed confounders, is consistent with the real confounders from the fifth and sixth panels of **Fig S7**. These results clearly validate the power of CIC to reconstruct confounders in a high-dimensional network.

**Causality inference and confounders reconstruction in real-world systems**
*Food chain network of plankton species*
We detected the causality in ecological network consisting of six creatures, i.e., Pico cyanobacteria, Rotifers, Nano flagellates, Calanoid copepods, Bacteria and Nutrients, where the first four species make up the observable food chain network because the latter two species are unobservable in this system. The oscillatory population data are selected from an 8-year mesocosm experiment of a plankton community isolated from the Baltic Sea (26-28), and the time series data with 794 time points were finally obtained after preprocessing. **Fig 3A** illustrates the true food chain network, in which Bacteria and Nutrients are two confounders without observation data. The second panel of **Fig 3A** shows that there are three confounders, i.e., Nano flagellates, Pico cyanobacteria and unobservable Bacteria for Rotifers and Calanoid copepods, and one other confounder Nutrients of Nano flagellates and Pico cyanobacteria. **Fig 3B** shows the identified causality of CIC and other five (2-variable) causal methods, in which the red grids represent wrong identifications. And the comparison with 3-variable models is shown in **Supplementary Text 2.3**. We can see that the true causalities in ground truth are all successfully detected by our method, while other methods, e.g., GC, TE, CCM, and CME provide the inferred networks containing more wrong causal links, which obviously violate the ecological rule. Among the compared methods, CMC performs relatively better but missed one causal link. In addition, **Fig 3C** provides a more obvious comparison for the performance of these methods in terms of ROC curve and Accuracy. In addition, our method also successfully identifies the confounders, including Nano flagellates, Pico cyanobacteria and Bacteria. However, the confounder effect caused by Nutrients is misidentified as a causal link, probably because of the strong noises of the real data. Moreover, we also quantify confounders in **Fig 3D**, where the correlation between reconstructed confounders and the true confounders, e.g., Nano flagellates, Pico cyanobacteria are above 0.25. This value is much lower than the correlations in previous systems, probably due to the unknown observation data of a primary true confounder, i.e., Bacteria. Fortunately, the reconstruction of confounders is effective compared to the non-confounder case, e.g., with the average value of 0.10. In summary, the CIC accurately inferred all causal links, identified partial confounders and effectively quantifies these confounders, in this real-world system.

*Hong Kong air pollution and cardiovascular disease*

Strong associations between individual air pollutants and cardiovascular diseases (CVDs) have been found in previous study (29). The data of concentrations of air pollutants and hospital admission of cardiovascular diseases in Hong Kong from 1994 to 1997 are used to infer the causality in this work. The result is shown in **Fig3E-H**. **Fig 3E** describes the true network among the pollutants and CVDs, $SO_2$, $O_3$, $NO_2$, respirable suspended particulates (RSP), and CVDs, and the second panel shows the ground truth. The first panel of **Fig 3F** shows the inference result of CIC, i.e., the RSP, SO2 are the cause of CVDs, which is consistent with previous studies (3, 31, 32). In addition, a unidirectional causal interaction from SO2 to NO2 is detected, which is also founded in previous study (1). However, the causality from NO2 to CVDs is not detected from the available data, as the causal link is regarded as the confounder effect of NO2 and CVDs by CIC. It cannot be excluded other factors which are not completely available, such as temperature, air humidity, etc. By contrast, other methods lead to at least four false positive links when setting the 65% quantile (3, 4) of all causality strengths between different nodes as the threshold value, such as, the causal links from CVDs to NO2, O3 by GC and the causal links from CVDs to RSP, SO2, NO2 and O3 by TE, which obviously violates the ecological rule. And other methods, e.g., CCM, CME and CMC cannot effectively infer the causality from pollutants to CVD, which may be affected by the strong noises of observed data or invisible confounders. And **Fig 3G** provides a quantitative comparison for these methods with (2-variable) causal methods by ROC curve and Accuracy, and the comparison with 3-variable models is shown in **Supplementary Text 2.3**. the result shows that our method outperforms other methods significantly by ROC curve and Accuracy. In addition, the confounders are also effectively identified by CIC, such as the confounder of SO2 and $O_3$, the confounder of CVDs and $O_3$ and the confounder of $NO_2$ and $O_3$. **Fig 3H** shows the identification and quantification of confounders, the correlation between the reconstructed one and the true confounders approaches to ~ 0.5, and the average correlation of the non-confounder case is 0.2, indicating the power of our method in revealing the causal mechanism of the system.

*Rat circadian rhythm gene expression data*

The circadian rhythm system is responsible for regulating various physiological and behavioral rhythms (33). Extensive studies have revealed that a set of key circadian genes are capable of generating circadian oscillations (30). However, how these circadian genes drive the circadian oscillation of gene expressions remain unknown. Thus, reconstructing the network of regulatory interactions among these circadian genes from a causal perspective is vital for understanding the processes that underlie circadian rhythms. Here, we used the CIC to a rat circadian rhythm gene expression data to detect causality among gene to reconstruct circadian network. The dataset includes four experiments (**Fig 4A**), i.e. one control and three phase-shifted experiments. We provided the detailes about the dataset in the **Supplementary Text 2.5**.

In **Fig 4B**, we illustrate the dynamics of gene expression for the seven key circadian genes in the four experiments. The circadian oscillation of *Per1* expression can be used to identify the immediately phase-shifted by forskolin, and the master clock gene *Perl* shows phase unchanged in the CT6 experiment, phase delay in the CT14 experiment and phase advance in the CT22 experiment (**Fig 4B**). Then, we use CIC algorithm to elucidate the gene regulatory network in the four different experiments. Here we consider a core regulatory network consisting of 18 genes, in which the transcriptional circuits are formed by regulation of E/E' boxes, DBP/E4BP4 binding elements, and RevErbA/ROR binding elements, respectively (34, 35). **Fig 4C** illustrate the ground truth of the 18-node circadian rhythm network. **Fig 4D** shows the reconstructed circadian network for the four stage, control, phase unchanged, phase delay and phase advance, and the hub networks of the seven key circadian genes, e.g. *Cry1*, *Cry2*, *Per1*, *Per2*, *Clock*, *Npas2* and *Bmal1*(*Arntl*), are provided in the top right of **Fig 4C**. It is clear that *Clock* and *Cry2* are two key genes that distinguish phase shifts. In the control and CT6 experiment, *Clock* gene is especially active to regulate other genes and *Cry2* is inactive with the lower degree. In the CT14 and CT22 experiments, the *Clock* gene become inactive with less causal links and the *Cry2* begin to play a core role in the network. Specially, at the CT6 experiments, after the stimulus by forskolin, the genes, *Per1*, *Per2* and *Npas2* actively participate in the regulation of the circadian rhythm network with dense causal links. In addition, for the phase delay stage in the CT14 experiment, the interactions of circadian rhythm network increase and the regulation effects of the hub network are denser than the control, such as the degree of *Per1* and *Per2* both increases from 4 (of control) to 8. For the phase advance in the CT22 experiment, regulations of the circadian network become weaker and there are very sparse association for the hub network. In addition, **Fig 4E** further shows the ROC curve of CIC and the other eight compared causal methods for the circadian network reconstructions, CIC performs better for the inference of real circadian rhythm network with a AUROC of 0.743. **Fig 4F-G** further demonstrates the potential of the method in identifying and reconstructing confounders. In **Fig 4F**, there is a gene *Bmal1* which points to genes *Clock* and *Dec2*, then we remove the gene *Bmal1* from the 18-node circadian network, i.e. it becomes an invisible confounder of *Clock* and *Dec2*. Then we calculated the CIC index to detect the causality between *Clock* and *Dec2*, yielding a CIC score of 0.6631. Then there are confounders of *Clock* and *Dec2* according to the Theorem 1. Next, the correlation between the common information of *Clock* and *Dec2* disentangled by CIC and the real confounder *Bmal1*, with a correlation of 0.3463. In addition, we also reconstruct other confounders of the network in the same way, as shown in **Fig 4G**, the average correlation between the reconstructed confounders and real confounders is 0.35, while for other nodes without confounders, their correlation between the node with the lowest degree in the network approaches to 0, indicating the quality of such reconstruction by CIC. The low correlation values of 0.35 between the reconstructed confounders and real confounders may be caused by other invisible confounders outside of the 18-node circadian network. In summary, these above results demonstrate the ability of our method in reconstructing circadian rhythm network.

**Discussion**

In this work, we developed a novel method, CIC, to infer dynamical causality, even in the presence of invisible confounders. Our approach builds upon novel theoretical results, i.e., Orthogonal Decomposition Theorem and its realization VAE framework from time-series data. Our theoretical foundation ensures the causal detection using only two observed variables (data) for any high-dimensional system even if there are many unobserved confounders, which is actually a long-standing problem in the field. In addition to the solid theoretical foundation, there are two unique advantages of this method: 1) detecting causal interactions with invisible confounders using the data of only two observed variables and 2) reconstructing the invisible confounders. Besides the invisible confounder problem, our CIC is able to handle another notorious non-separability problem in causal inference.

Generally, the dynamics of variables in a nonlinear system are regarded non-separable due to their intertwined interactions. Although many approaches proposed recently endeavor to inferring causality, such as prediction-based methods and cross-mapping-based methods, they either misidentify the common interactions from confounders as causality or loss accuracy when a large number of confounders are in the system. A virtue of our method is that it can distinguish the relationships between variables, e.g., the actual causality or the common interaction caused by invisible confounders, based on the observations of pairwise variables solely. This eliminates the need to introduce other observed variables like the condition-based approaches. The central idea lies in the orthogonal decomposition theorem in a delay embedding space, which theoretically and computationally transforms non-separable dynamics of intertwined variables in the original space into their separable dynamics in the embedding space, thus also solving the non-separability problem of causal inference. In addition, to distinguish ddirect and indirect causality and identify visible and invisible confounders, we extend Theorem 1 to a conditional version involving multiple observed variables, as outlined in Theorem 2 of Supplementary Materials. In addition, computationally we adopt deep neural networks to efficiently realize the orthogonal decomposition so as to reliably detect causal links merely from observed time-series data even with unobserved confounders. The method is also applied to various benchmark and real-world datasets, demonstrating the power for inferring causality and invisible confounders not only in a low-dimension system but also in a high-dimension system. Explicitly inferring the causalities and confounders is the key to recover the causal mechanisms of a complex system. Therefore, the CIC provides a powerful tool along this feat and effectively solves the confounder problem arising from previous influential research.

It is worth noting that there remain several issues that merit further investigation. Firstly, CIC framework mainly relies on the temporal information of the observed or measured data, thus the sample size of time series should not be too small (e.g., $\geq 10$) to maintain the algorithm performance. Secondly, identifying confounders between x and y requires training two VAE models: one from x to y and another from y to x. Investigating how to identify confounders using this method is worthwhile. Thirdly, our study primarily focuses on dynamical causality within a non-intervention system. Addressing how to infer dynamical causality with interventions based solely on observed time series remains an important and intriguing problem for future research.

**Materials and Methods**

**Overview of CIC method**

A real system is generally composed of a large number of variables, many of which are unobservable/invisible but affect the causal inference between variables, knowns as invisible confounders. The invisible confounders have brought up two challenges for causality detection from observed data:

1) How to infer the causality from $x$ to $y$ in the presence of invisible confounders?
2) How to reconstruct those invisible confounders?

To solve such problems, a new framework based on orthogonal decomposition theorem in a delay embedding space is proposed to infer causality even under invisible confounders (**Fig 5A**). Assume that time series (length *L*) of two variables $x_t$ and $y_t$, e.g., $\{x_1, \ldots, x_t, \ldots x_L\}$ and $\{y_1, \ldots, y_t, \ldots y_L\}$ are observed in the original space of a dynamical system, while all other variables in the system are not observed. The causality in the original space is defined as follows: if $y_t = f(x_{t-1}, y_{t-1})$, $x_t = g(x_{t-1})$, then $x \to y$. However, there is a case that will be misidentified as causality, i.e., $y_t = f(y_{t-1}, z_{t-1})$, $x_t = g(x_{t-1}, z_{t-1})$, in which $z_{t-1}$ is an invisible confounder of $x_t$ and $y_t$. Then only with observed data of $x$ and $y$, it is difficult to infer correct causality $x \to y$ in the original space due to their invisible confounder set *z* and the non-separability of $x_t$ and $y_t$.

To overcome these problems, the Takens' delay embedding theory is introduced to transform the original data $x$ and $y$ into the embedding data $X$ and $Y$, where $X_t = [x_t, x_{t-1}, \ldots, x_{t-p}]^T$ and $Y_t = [y_t, y_{t-1}, \ldots, y_{t-p}]^T$, respectively, i.e. $X_t, Y_t \in R^{p+1}$ (**Fig 5B**). Then the embedding space $X_{t-1} = \{z_x, z_{xy} \mid z_x \perp z_{xy}\}$ or $Y_t = \{z_y, z_{xy} \mid z_y \perp z_{xy}\}$ is theoretically ensured to decompose into two orthogonal subspaces at each instant *t* respectively, i.e., the common space/information $z_{xy}$ between $X_{t-1}$ and $Y_t$, and the private space/information $z_x$ (or $z_y$). Thus, clearly $z_x \perp Y_t$ and $z_y \perp X_{t-1}$. In this paper, we drop *t* from $z_x, z_y, z_{xy}$ for the sake of simplicity. Then, according to our orthogonal decomposition theorem, if $||z_x|| = 0$ or $X_{t-1} = F(z_{xy})$ indicating that all information of $X_{t-1}$ is contained in $Y_t$, then $x \to y$. Otherwise, if $||z_x|| \neq 0$ or $X_{t-1} \neq F(z_{xy})$, then $x \not\to y$. Thus, we can construct the causal criterion under invisible confounders, i.e. $\text{CIC}_{x \to y}$.

Here the deep learning techniques are introduced as the backbone to computationally realize such information decomposition of variables, as shown in **Fig 5C**. The input of our CIC (causality under invisible confounders) method is the time-series in an embedding space transformed from the observed time-series in the original space by the delay-embedding theory. Then the embedded variables' information was encoded into two orthogonal latent subspaces, i.e., private latent subspace and shared latent subspace. The output of CIC is the reconstructed embedding space data, which is used to calculate the causal index $\text{CIC}_{x \to y}$ (**Fig 5C**). And the invisible confounder can be quantified/reconstructed by the shared latent subspace (**Fig 5D**). In summary, our CIC serves two key functions: firstly, inferring the causality between pairwise variables from the observed data, and secondly, reconstructing the invisible confounders of pairwise variables.

**Orthogonal decomposition theorem for dynamical causality**

Consider a general discrete dynamical system with *n* variables in the form of
$$\boldsymbol{x}_t = \boldsymbol{f}(\boldsymbol{x}_{t-1}), \tag{1.1}$$
with an initial value $x_0$ and $t = 1, 2, \ldots$, where $\boldsymbol{f} = [f_1, f_2, \ldots, f_n]^T$ is a nonlinear function vector, and $\boldsymbol{x}_t = [x_{1,t}, x_{2,t}, \ldots, x_{n,t}]^T$ is the system's state vector which describes the evolution of the system along time. For the case of two observed variables $(x, y)$ shown in Fig.5, we simply have $x_{i,t} = x_t$ and $x_{j,t} = y_t$. Here, one unit time interval is $\Delta t$. We first define dynamical causality in the original space.

**Definition 1 (Dynamical causality defined in original space)**: We call that $x_i$ has the dynamical causality (DC) to $x_j$, or the dynamics of $x_j$ depend on $x_i$, if the following condition of eq. (1.1) holds

$$\frac{\partial f_j(x_{t-1})}{\partial x_{i,t-1}} \neq 0$$

for almost any sampled $t$, and we denote it as $x_i \to x_j$. Otherwise, we say that $x_i$ has no dynamical causality to $x_j$, i.e., $x_i \nrightarrow x_j$ (3, 4).

We next transform the original space to a delay embedding space. For the sake of simplicity in description, we consider a discrete dynamical system of $x \to y$ with $z$ as an invisible variable (or vector), as follows

$$\begin{cases} x_t = g(x_{t-1}; z_{t-1}), \\ y_t = f(x_{t-1}, y_{t-1}; z_{t-1}), \end{cases} \quad (1.2)$$

where $f$ and $g$ are two nonlinear functions, i.e. $x_t = x_{i,t}$ and $y_t = x_{j,t}$ with $z_t = x_{k,t}$ for eq. (1.1). By combining all $p$ equations of $f(x_{t-1}, y_{t-1}; z_{t-1}) - y_t = 0$ with $t, t-1, \ldots, t-(p-1)$ of the second eq. (1.2), we have

$$H(X_{t-1}, Y_t; Z_{t-1}) = 0, \quad (1.3)$$

where $H = [f(x_{t-1}, y_{t-1}; z_{t-1}) - y_t, \ldots, f(x_{t-p}, y_{t-p}; z_{t-1}) - y_{t-(p-1)}]^T$ is a nonlinear function vector; $Y_t = (y_t, y_{t-1}, \ldots, y_{t-p})^T \in R^{p+1}$ and $X_{t-1} = (x_{t-1}, x_{t-2}, \ldots, x_{t-p})^T \in R^p$ with $Z_{t-1} = (z_{t-1}, z_{t-2}, \ldots, z_{t-p})^T \in R^p$ are the delay embedding vectors of variables $x$ and $y$ with $z$, respectively. Clearly, $\frac{\partial f(x_{t-1}, y_{t-1}; z_{t-1})}{\partial x_{t-1}} \neq 0$ for almost any sampled $t$ due to $x \to y$ or Definition 1, and thus we have the $p \times p$ diagonal matrix $|\frac{\partial H(X_{t-1}, Y_t; Z_{t-1})}{\partial X_{t-1}}| \neq 0$. Hence, we can apply the implicit function theorem to eq. (1.3) to derive $X_{t-1}$ as

$$X_{t-1} = F(Y_t; Z_{t-1}) = F(Y_t), \quad (1.4)$$

where the last term is used due to invisible $z$ to simply the description. And the derivation of eq. (1.4) based on a detailed version with noise term of system eq. (1.2) or eq. (1.1) is provided in **Supplementary Materials 1.1**. Furthermore, Taken's embedding theorem also ensures the function $F$ of eq. (1.4) when the embedding dimension $p > 2d$, where $d$ is the intrinsic dimension of attraction manifold of the dynamics for eq. (1.2). Generally, $d$ is small for a real system and we can choose an appropriate embedding dimension $p$. Note that we next define $X_{t-1} = (x_{t-1}, x_{t-2}, \ldots, x_{t-(p+1)})^T \in R^{p+1}$ and $Z_{t-1} = (z_{t-1}, z_{t-2}, \ldots, z_{t-(p+1)})^T \in R^{p+1}$ in eq.(1.4) for the simplicity of notation without loss of generality. Thus, we have the dynamical causality measured in the embedding space as follows.

**Definition 2 (Dynamical causality defined in delay embedding space)**: we call the dynamical causality (DC) from $x$ to $y$, i.e. $x \to y$, if there is the inverse reconstruction or inverse mapping $X_{t-1} = F(Y_t)$ in the delay embedding space $(X, Y)$ for eq. (1.2).

Note that the cause $X_{t-1}$ is the information at past time $t-1$, while its effect $Y_t$ is the information at current time $t$. We can reconstruct the past cause by the current effect if $x \to y$ in terms of the delay embedding data. For a system (1.1) with multiple observed variables, without loss of generality, we can similarly derive the direct dynamical causality $x_i \to x_j$ on the condition of all other observed variables, yielding the similar result

$$X_{t-1} = F(Y_t; W_{t-1}), \quad (1.5)$$

where $X_{t-1} = (x_{i,t-1}, \ldots, x_{i,t-p})^T, Y_t = (x_{j,t}, \ldots, x_{j,t-(p-1)})^T$, with $W_{t-1} = (x_{k,t-1}, \ldots, x_{k,t-(p-1)})^T$ as

the delay embedding vector of the all remaining observed variables except $x_i$ and $x_j$ in eq. (1.1). The definition of such direct dynamical causality for eq. (1.1) is given in **Supplementary Materials 1.1** based on eq. (1.5).

For any pair of observed variables defined in eq. (1.1) or (1.2), e.g. $x$ and $y$, we can perform an orthogonal decomposition for their embedding vectors $X_{t-1}$ and $Y_t$, respectively,

$$X_{t-1} = \{z_{x_{t-1}}, z_{xy_{t-1}} \mid z_{x_{t-1}} \perp z_{xy_{t-1}}\} \text{ and } Y_t = \{z_{y_t}, z_{xy_{t-1}} \mid z_{y_t} \perp z_{xy_{t-1}}\},$$

in which $\perp$ is the symbol of 'orthogonal' or independence, and $z_{xy_{t-1}}$ is the shared space between $X_{t-1}$ and $Y_t$. Then, we can easily show $z_{x_{t-1}} \perp Y_t$ and $z_{y_t} \perp X_{t-1}$. For clarity, we will omit the subscript $t$ or $t-1$ of decomposed variables in the following description. Consequently, according to Definition 2, we have the following theorem of an orthogonal decomposition in the latent space.

**Theorem 1 (Orthogonal decomposition theorem in delay embedding space):** Given time series data for any pair of the observed variables $x_t$ and $y_t$ defined in eq. (1.2), there exist appropriate injective nonlinear decomposition mappings $\mathcal{G}_x$ and $\mathcal{G}_y$ to orthogonally decompose the embedding vectors $X_{t-1}$ and $Y_t$, respectively, i.e.

$$\mathcal{G}_x(X_{t-1}) = \{z_x, z_{xy} \mid z_x \perp z_{xy}\} \text{ and } \mathcal{G}_y(Y_t) = \{z_y, z_{xy} \mid z_y \perp z_{xy}\}, \quad (1.6)$$

in which $\perp$ is the symbol of 'orthogonal' or independence. Furthermore, we have the following criteria of dynamical causality.
*Criterion 1* (Causality): $X_{t-1} = F(z_{xy})$ or $\|z_x\| = 0 \Leftrightarrow x \to y$;
*Criterion 2* (Non-causality): $X_{t-1} \neq F(z_{xy})$ or $\|z_x\| \neq 0 \Leftrightarrow x \not\to y$.

**Remark 1 (Confounder reconstruction):** For the case of non-causality from $x$ to $y$ of eq. (1.2), i.e. $X_{t-1} \neq F(z_{xy})$ or $\|z_x\| \neq 0$, we have the following criterion.
*Criterion 3* (Confounder): If $\|z_{xy}\| \neq 0$, then there exists a confounder of $x$ and $y$, which can be quantified by $z_{xy}$.

**Remark 2 (Causal index):** For any pair of observed variables defined in eq. (1.1) or (1.2), e.g. $x$ and $y$, we can quantify the dynamical causality from $x$ to $y$ as $\text{CIC}_{x \to y} = \frac{\|z_{xy}\|}{\|z_x\| + \|z_{xy}\|}$. If $\text{CIC}_{x \to y} = 0$, there is no causality from $x$ to $y$; if $\text{CIC}_{x \to y} = 1$, there is causality from $x$ to $y$; if $\text{CIC}_{x \to y} \in (0,1)$, there is a confounder of $x$ and $y$.

Here, $\|\cdot\|$ is an appropriate norm. The detailed proof of the three criteria and the Orthogonal Decomposition Theorem under multiple observed variables (to infer direct dynamical causality) are provided in the **Supplementary Materials 1.1**. In particular, Theorem 1 indicates that we can theoretically infer their causal relation using the time-series data of only two observed variables for any high-dimensional system, even if there are possibly many unobserved confounders, which is actually a long-standing problem in the field. To achieve this computationally, we adopt the Variational Auto Encoder (VAE) for nonlinear orthogonal decomposition, allowing form causality inference from the observed time series data (see **Supplementary Materials 1.2** for the details**)**.

**Computational framework for causality inferrence under invisible confounders by VAE**

To infer causality between pairs of variables, e.g. $x$ and $y$, in a system (1.1) or (1.2), firstly, the original time series data of $x_t$ and $y_t$ are transformed into delay embedding data $X_t$ and $Y_t$, respectively, according to the Takens' embedding theorem (**Fig 5B**). Then, we realize the orthogonal decomposition of variables and the inference of causal associations by a VAE framework, as shown in **Fig 5C**. The CIC takes the delayed embedding vectors of the variables $x$ and $y$ as input, i.e., $X_{t-1}$ and $Y_t$.

In the inference process, the delay embedded variable $X_{t-1}$ or $Y_t$ is encoded into two separate latent subspaces, i.e., private latent space $z_x$ or $z_y$ and common latent space $z_{xy}$ (i.e. $z_{xy} = z_{xy}^x = z_{xy}^y$) by two different encoders, $E_x$ parametrized by $\emptyset$ and $E_y$ parameterized by $\emptyset'$, respectively. Supposing that the private information or common information with $Y_t$ of $X_{t-1}$ is continuous, then for each sample, we sample the multivariate Gaussian

$$q_\emptyset(z_x|X_{t-1}) \sim \mathcal{N}\left(\mu_{z_x}, (\sigma_{z_x})^2\right), \tag{2.1}$$

$$q_\emptyset(z_{xy}^x|X_{t-1}) \sim \mathcal{N}\left(\mu_{z_{xy}^x}, (\sigma_{z_{xy}^x})^2\right), \tag{2.2}$$

where the means and standard deviations, i.e., $\{\mu_{z_x}, z_{xy}^x\}$ and $\{\mu_{z_{xy}^x}, \sigma_{z_{xy}^x}\}$ are generated by the reparameterization and split of the output of encoder $E_x$. Likewise, the disentangled representations of $Y_t$ are sampled from the multivariate Gaussian

$$q_{\emptyset'}(z_y|Y_t) \sim \mathcal{N}\left(\mu_{z_y}, (\sigma_{z_y})^2\right), \tag{2.3}$$

$$q_{\emptyset'}(z_{xy}^y|Y_t) \sim \mathcal{N}(\mu_{z_{xy}^y}, (\sigma_{z_{xy}^y})^2). \tag{2.4}$$

In the generative process, the concatenation of the common latent vector $z_{xy}^x$ and the private latent vector $z_x$ are expected to reconstruct the original delay embedding vector $X_{t-1}$ through the Decoder $D_x$ parametrized by $\theta$, i.e., $\hat{X}_{t-1}$. And the concatenation of the latent spaces $z_{xy}^y$ and $z_y$ are expected to reconstruct the original delay embedding vector $Y_t$ through the Decoder $D_y$ parametrized by $\theta'$, i.e., $\hat{Y}_t$. Note that there are also two private/special reconstructions by $z_x$ and $z_{xy}$, respectively (see Fig S3). Here, $\hat{X}_{t-1}^{z_{xy}}$ is the reconstruction of $X_{t-1}$ by the common latent vector $z_{xy}$ and an all-zero vector $z_x^0$ with the same dimension like $z_x$, which can also be regarded as a reconstruction of $X_{t-1}$ by $Y_t$, and $\hat{X}_{t-1}^{z_x}$ is the reconstruction of $X_{t-1}$ by the private latent vector $z_x$ and an all-zero vector $z_{xy}^0$ with the same dimension of $z_{xy}$.

The variational inference aims to maximize the evidence lower bound (ELBO) by learning the encoder $E_x$ (or $E_y$) and the decoder $D_x$ (or $D_y$), i.e.,

$$\mathcal{L}_{VAE} = \alpha E_{q_\emptyset(z_x,z_{xy}^x|X_{t-1})}[\log p_\theta(X_{t-1}|z_x,z_{xy}^x)] + \alpha E_{q_{\emptyset'}(z_y,z_{xy}^y|Y_t)}[\log p_{\theta'}(Y_t|z_y,z_{xy}^y)] - D_{KL}(q||p),$$

where $\alpha$ is a weight paramter, and $D_{KL}(q||p)$ is the Kullback-Leibeler divergence regularization term which regulates the variational posterior to follow the prior distribution. In addition, some constraints on the latent spaces are also introduced to control the training process. First, according to Theorem 1, the orthogonality constraint, i.e., $\mathcal{L}_{diff}$ is introduced as follows,

$$\mathcal{L}_{diff} = ortho(z_x, z_{xy}^x) + ortho(z_y, z_{xy}^y) + ortho(z_x, z_y),$$

and $ortho$ is defined by cosine similarity as follows,

$$ortho(X,Y) = \|cos(X,Y)\|^2 = \frac{\|X \cdot Y\|^2}{\|X\|^2 \|Y\|^2}.$$

Second, the sharing information $z_{xy}^x$ from $X_{t-1}$ and $z_{xy}^y$ from $Y_t$ are equivalent, i.e., $z_{xy} = z_{xy}^x = z_{xy}^y$, thus a soft constraint of equivalence is introduced into the model,

$$\mathcal{L}_{equal} = MSE(z_{xy}^x, z_{xy}^y).$$

Finally, the objective of CIC is to minimize the total loss, which is the weighted average of the losses $\mathcal{L}_{VAE}$, $\mathcal{L}_{diff}$ and $\mathcal{L}_{equal}$, i.e.,

$$\mathcal{L}_{loss} = \mathcal{L}_{VAE} + \beta_1 \mathcal{L}_{diff} + \beta_2 \mathcal{L}_{equal}, \tag{2.5}$$

where $\beta_1$ is the orthogonal loss coefficient and $\beta_2$ is the equivalent loss coefficient. A detailed version of the algorithm and implementation details are provided in the **Supplementary Materials 1.3**.

After the training of CIC with delay embedding vectors, the two latent spaces of $X$, i.e., $z_{xy}$ and $z_x$ generated by the encoder $E_x$, can be used to infer the causality from $x$ to $y$ even under invisible confounders according to Theorem 1 (Fig S3). Note $CIC_{x \to y} \in [0,1]$. Generally, CIC index is between 0 and 1 due to large amounts of noise and distractions contained in real-world data. Therefore, thresholds, i.e. $m$ and $M$, are selected to improve the robustness of CIC (in this work, $m$ and $M$ are set to 0.25 and 0.75 respectively). Consequently, the CIC can be used to infer causality in the following way:

(1) if $CIC_{x \to y} \in [0, m]$, then $x \nrightarrow y$;
(2) if $CIC_{x \to y} \in (m, M)$, then $x \nrightarrow y$ with confounder;
(3) if $CIC_{x \to y} \in [M, 1]$, then $x \to y$.

The robustness and stability of the causal index were further validated under different perturbed parameters. As shown in **Fig 5D**, the CIC score initially exhibited fluctuations but eventually stabilized to an asymptotic value as the perturbation parameters change. In addition, when $CIC_{x \to y} \geq M$, the higher the CIC score, the stronger the causality. In addition, in the CIC framework, the invisible confounders can be quantified, i.e., as shown in **Fig 5D**, the common information $z_{xy}$ decomposed from $x$ and $y$ represents the invisible confounder of $x$ and $y$. To validate the consistency between the hidden variable $z_{xy}$ and the real confounder $Z$, the Canonical Correlation Analysis (CCA) was used to estimate the correlation between $z_{xy}$ and $Z$, which can solve the problem of dimensional mismatch between hidden variables and confounders. The calculation details are as follows,

$$CCA(z_{xy}, z) = \sum_{i=1,j=1}^{n,m} \frac{Cov(z_{xy_i}, Z_j)}{\sqrt{Var(z_{xy_i})Var(Z_j)}},$$

where the $n$ and $m$ are the number of the samples of $z_{xy}$ and $z$, respectively. $z_{xy_i}$ is the $i$-th element of $z_{xy}$, $z_j$ is the $j$-th element of $z$.

**Acknowledgments**

This work has been supported by National Key R&D Program of China (2022YFA1004800), Strategic Priority Research Program of the Chinese Academy of Sciences (XDB38040400), Natural Science Foundation of China (62173271, 61873202, 31930022, 12131020, T2341007, T2350003, 12301620), Science and Technology Commission of Shanghai Municipality (23JS1401300), and JST Moonshot R&D(JPMJMS2021).

# Figures and Tables

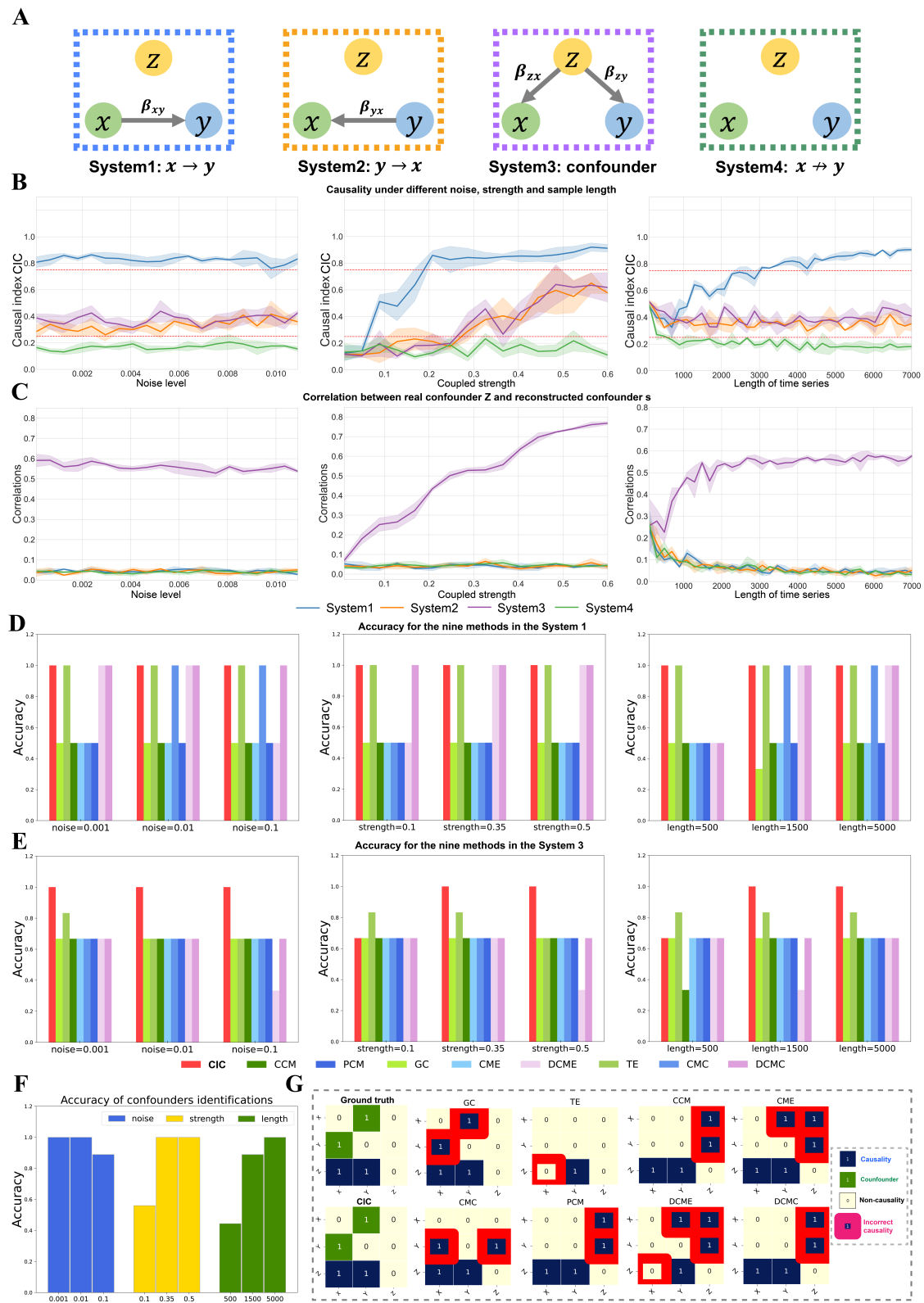

**Figure 1. Detecting causality in the three-dimensional system with different couplings.**
**A.** Four different logistic systems. The associations between $x$ and $y$ under the four cases are illustrated, i.e., System1: the causality from $x$ to $y$ ($\beta_{xy} = 0.35$, other coupled strengths=0),

System2: the inverse causality from $y$ to $x$ ($\beta_{yx} = 0.35$), System3: the confounder of $x$ and $y$ ($\beta_{zx} = 0.35, \beta_{zy} = 0.35$), and System4: non-causality case (all the coupled strengths=0). **B.** The robustness and effectiveness of CIC under different parameters, e.g. noise level, coupling strength and length of time series, are provided. **C.** The reconstruction or quantification for invisible confounders by CCA correlations under different parameters. **D.** The comparison with other eight approaches for the System1, and the accuracy of the nine methods are shown in the histogram. **E.** The comparison with other eight approaches for the System3. **F.** The accuracy of CIC in identifying confounders under different parameters. **G.** The detailed comparisons between CIC and other methods with the noise, causal strength, and length as 0.001, 0.35, and 5000, respectively. The first heat map is the ground truth of System3, where the horizontal coordinate is the cause and the vertical coordinate is the effect. The other heat maps are the identifications of CIC and other eight methods. The blue grids represent the causality, the green grids represent the confounders, and the yellow grids represent non-causality case. The red grids represent the misidentification.

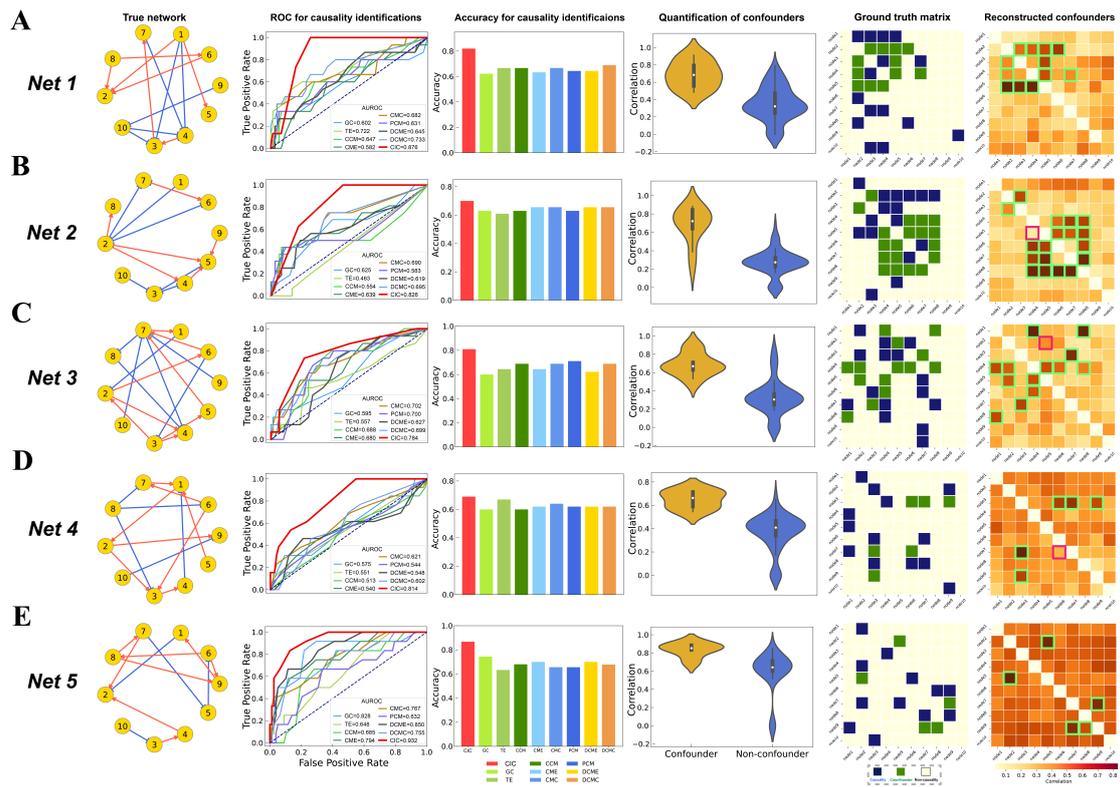

**Figure 2. Detecting causality in gene regulatory network from DREAM4. A-E.** The results of CIC in five 10-node networks. The first panel shows the true networks. The subsequent panels, from left to right, depict the comparative analysis between CIC and other methods using ROC curves, followed by bar plots representing accuracy metrics. The fourth panel presents the global correlation between the reconstructed confounders (common latent space) and the actual confounders, which approaches 0.7. The correlation between the common latent space for non-confounder cases and the node with the lowest degree is less than 0.3. The fifth collum illustrates the causalities and confounders in the ground truth matrix, in which the blue grids represent the causality, the green grids represent the confounders, and the sixth panel shows the reconstructions of confounders by our method.

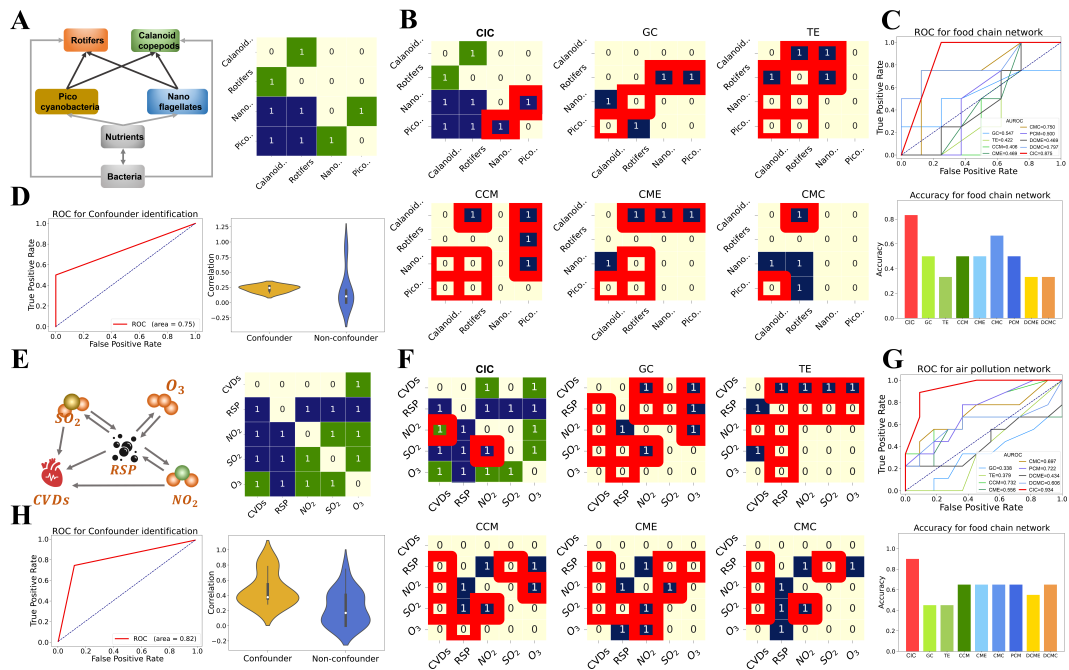

**Figure 3.** Detecting causal links in two real-world networks. Food chain network of plankton species: A. The true food chain network, in which Bacteria and Nutrients are two confounders without observation data. The second graph is the ground truth matrix, where the blue and green grids represent the true causality and confounder. B. The identifications of CIC and other five (2-variable) causal methods, in which the red grids represent erroneous identifications. C. The comparison of these methods by ROC curve and Accuracy. D. The identification and quantification of confounders in food chain network. Hong Kong air pollution and cardiovascular disease: E. The first graph is the true network among the pollutants, SO2, O3, NO2, respirable suspended particulates (RSP), and CVDs, and the second graph shows the ground truth in which the blue grids and green grids represent the true causality and confounders of the left network, respectively. F. The identifications of CIC and other five (2-variable) methods in air pollution and cardiovascular disease network. G. The comparison of these methods based on ROC curve and Accuracy. H. The identification and quantification of confounders.

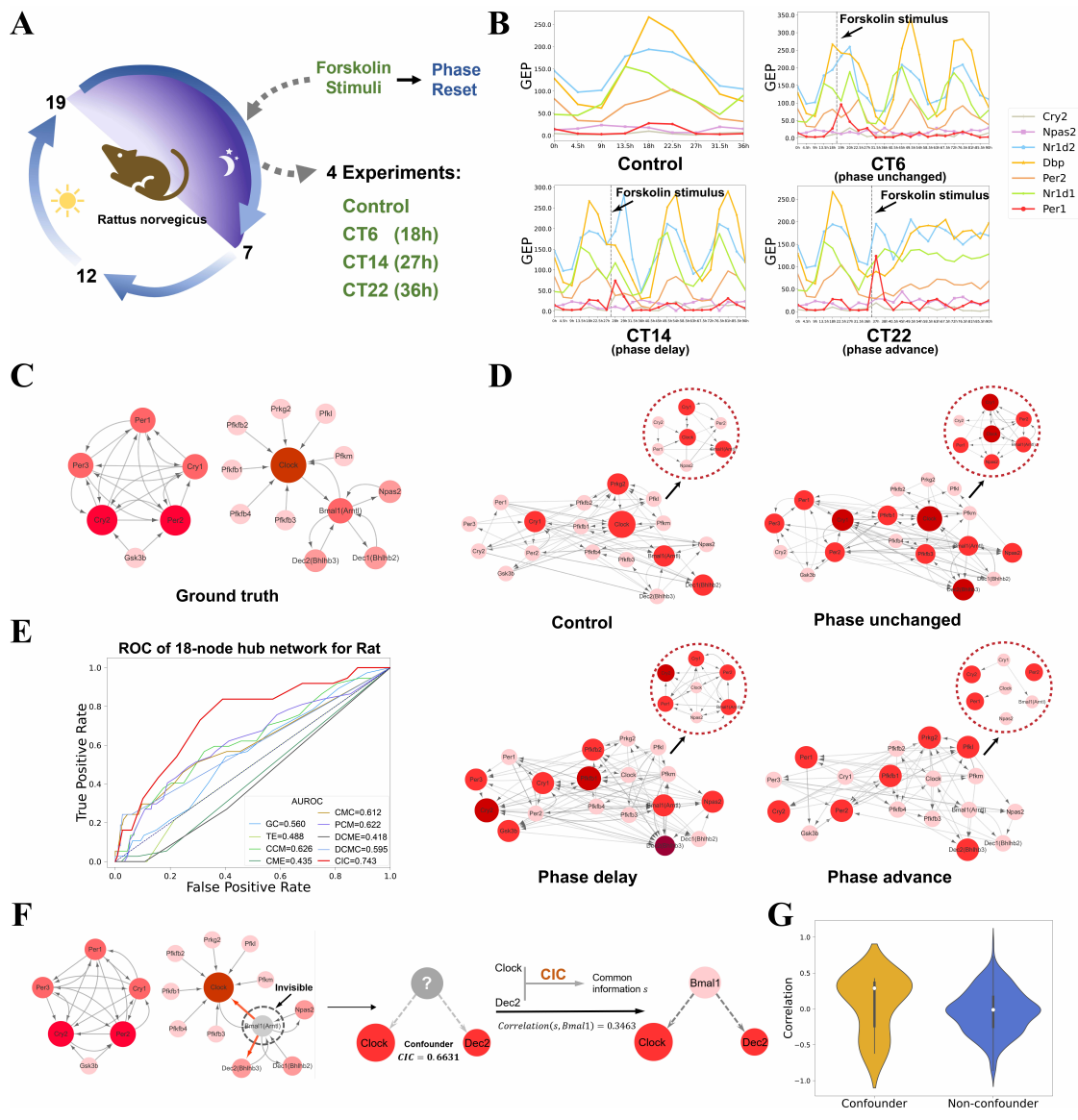

**Figure 4. Reconstructing the circadian rhythm network of rat. A.** Schematic diagram of the circadian rhythm experiments. **B.** The oscillation of several key circadian genes in the four experiments. **C.** The ground truth of the 18-gene circadian rhythm network. **D.** The inferred circadian network for the four stage, control, phase unchanged, phase delay and phase advance, and the hub networks of the seven key circadian genes are provided in the top right. **E.** The ROC curve of CIC with other eight methods for the 18-gene circadian network of control experiment. **F.** Since a gene *Bmal1* points to gene *Clock* and *Dec2*, the gene *Bmal1* from the 18-gene circadian network is removed. And the CIC is used to reconstruct the confounders between *Clock* and *Dec2*. **G.** the performance of our method in quantifying confounders of the circadian rhythm network.

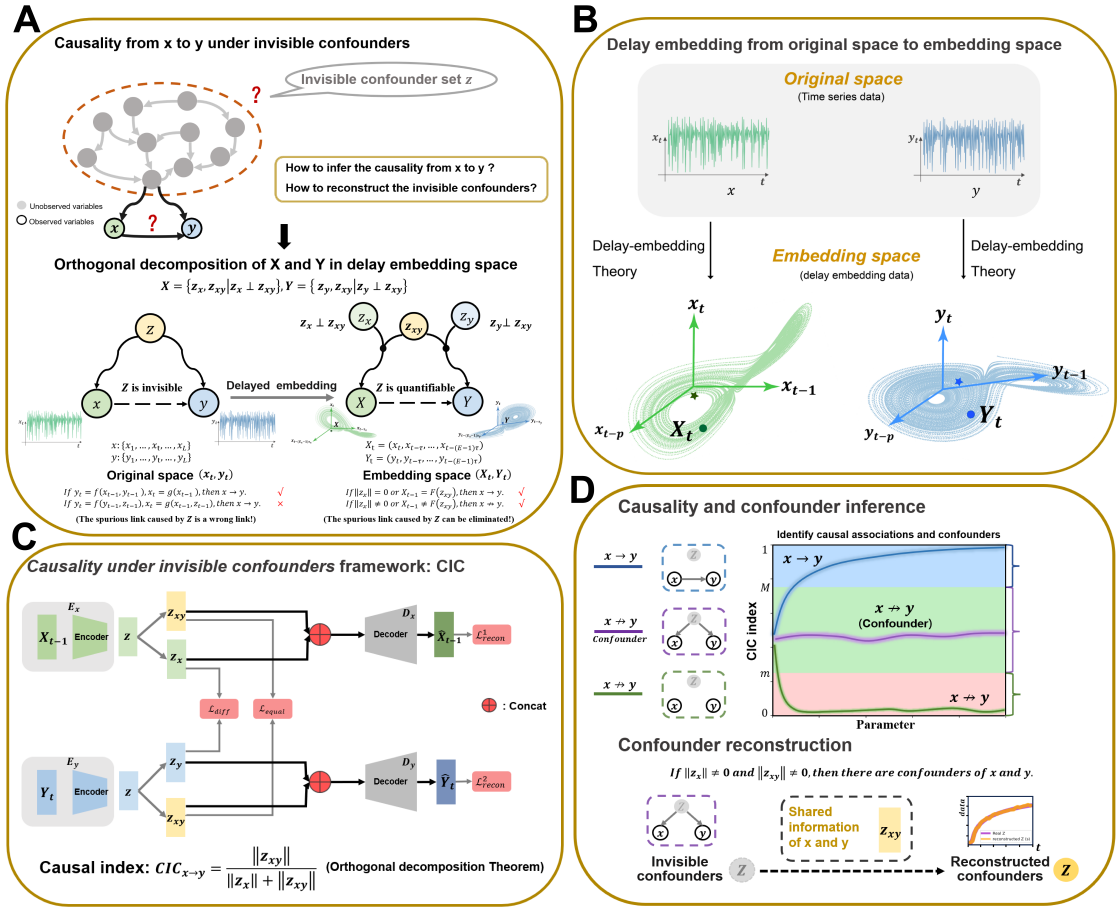

**Figure 5. Framework of the Causality under Invisible Confounders (CIC) method. A. Basic principles of CIC.** Complex dynamical systems contain many unobservable variables affecting the causal inference between variables, which brings up two challenges. To solve the problems, a CIC framework based on Orthogonal Decomposition Theorem is proposed to eliminate the spurious links that are misidentified as causal ones in original space. **B. Delay embedding space of $x$ and $y$.** Time series of original variables $x_t$ and $y_t$ can be transformed into those of delay embedded variables $X_{t-1}$ and $Y_t$. **C. CIC framework.** The delay embedded variable $X_{t-1}$ or $Y_t$ are encoded into two orthogonal latent subspaces, private subspace $z_x$ (or $z_y$) and common subspace $z_{xy}$. The decoders reconstruct the inputs through the concatenation of private and common subspaces. Concat means concatenation; $\mathcal{L}_{diff}$ and $\mathcal{L}_{equal}$ are the orthogonality and equivalent constraint, respectively. And the two subspaces $\|z_x\|$ and $\|z_{xy}\|$ are used to construct the causal index $\text{CIC}_{x \to y}$. **D. Detecting the causality and reconstructing invisible confounder**. The blue curve, purple curve, and green curve indicate the causality, a confounder, and the non-causality of *x* and *y*, respectively. The invisible confounder Z of x and y can be quantified through CIC framework.